\documentclass[sigconf]{acmart}

\usepackage{booktabs} % For formal tables
\usepackage{amssymb}
\usepackage{xspace}
\usepackage{multirow}
\usepackage{flushend}

\copyrightyear{2017}
\acmYear{2017}
\setcopyright{licensedothergov}
\acmConference{ICMR '17}{June 06-09, 2017}{Bucharest, Romania}\acmPrice{15.00}\acmDOI{http://dx.doi.org/10.1145/3078971.3079033}
\acmISBN{978-1-4503-4701-3/17/06}

\def\ie{\textit{i.e.}~}
\def\eg{\textit{e.g.}~}
\def\pp{pan2pan\xspace}
\def\pps{\pp/sum\xspace}
\def\ppp{\pp/pinv\xspace}
\def\ppn{\pp/net\xspace}
\def\ii{im2im\xspace}
\def\ip{im2pan\xspace}
\def\pi{pan2im\xspace}

\usepackage{pgfplots}
\usepackage{pgfplotstable}
\usepackage{pgf,tikz}
\usetikzlibrary{arrows,shapes,calc,matrix,fit}

\usepgfplotslibrary{external}
\newcommand{\extfig}[2]{\tikzsetnextfilename{figs/extern/#1}{#2}}
\newcommand{\extdata}[1]{\input{#1}}
\newcommand{\leg}[1]{\addlegendentry{#1}}

\begin{document}
\title{Panorama to panorama matching for location recognition}
%\titlenote{Produces the permission block, and
%  copyright information}
%\subtitle{Extended Abstract}
%\subtitlenote{The full version of the author's guide is available as
%  \texttt{acmart.pdf} document}

\author{\mbox{Ahmet Iscen$^1$ \hspace{10pt} Giorgos Tolias$^2$ \hspace{10pt} Yannis Avrithis$^1$ \hspace{10pt} Teddy Furon$^1$ \hspace{10pt} Ond{\v r}ej Chum$^2$}}
\affiliation{$^1$ Inria, Rennes, France}
\affiliation{$^2$ Visual Recognition Group, Faculty of Electrical Engineering, CTU in Prague, Czech Republic}
\email{}

\renewcommand{\authors}{A. Iscen, G. Tolias, Y. Avrithis, T. Furon and O. Chum}
\renewcommand{\shortauthors}{\authors}

% \author{Ahmet Iscen}
% \affiliation{%
%   \institution{Inria} 
%   \city{Rennes} 
%   \state{France} 
% }
% \email{ahmet.iscen@inria.fr}

% \author{Giorgos Tolias}
% \affiliation{%
%   \institution{VRG, FEE, CTU}
%   \city{Prague} 
%   \state{Czech Republic} 
% }
% \email{giorgos.tolias@cmp.felk.cvut.cz}

% \author{Yannis Avrithis}
% \affiliation{%
%   \institution{Inria}
%   \city{Rennes} 
%   \state{France} 
% }
% \email{ioannis.avrithis@inria.fr}

% \author{Teddy Furon}
% \affiliation{%
%   \institution{Inria}
%   \city{Rennes} 
%   \state{France} 
% }
% \email{teddy.furon@inria.fr}

% \author{Ond{\v r}ej Chum}
% \affiliation{%
%   \institution{VRG, FEE, CTU}
%   \city{Prague} 
%   \state{Czech Republic} 
% }
% \email{chum@cmp.felk.cvut.cz}

\begin{abstract}
Location recognition is commonly treated as visual instance retrieval on ``street view'' imagery. The dataset items and queries are panoramic views, \ie groups of images taken at a single location. This work introduces a novel panorama-to-panorama matching process, either by aggregating features of individual images in a group or by explicitly constructing a larger panorama. In either case, multiple views are used as queries. We reach near perfect location recognition on a standard benchmark with only four query views.
\end{abstract}



%
%% We no longer use \terms command
%%\terms{Theory}
%

\maketitle

\def\etal{\textrm{et al}.\,}
\def \X{\mathbf{X}}
\def \x{\mathbf{x}}
\def \Y{\mathbf{Y}}
\def \y{\mathbf{y}}
\def \1{\mathbf{1}}
\def \G{\mathbf{G}}
\def \vv{\mathbf{v}}
\def\ie{\emph{i.e.}~}

\newcommand{\real}{\ensuremath{{R}}\xspace}

\newcommand{\mypar}[1]{\noindent \textbf{#1}}

\newcommand{\alert}[1]{{\color{red}{#1}}}

\def\sssp{\hspace{1pt}}
\def\ssp{\hspace{3pt}}
\def\msp{\hspace{5pt}}
\def\bsp{\hspace{12pt}}

\section{Introduction}
Location recognition has been treated as a visual instance retrieval task for many years~\cite{ZhKo06,ScBS07,ArZi14,AGT+15,ΗΕ08,SHSP16}. Additional, task-specific approaches include ground truth locations to find informative features~\cite{ScBS07}, regression for a more precise localization~\cite{ToSP11,KGC15}, or representation of the dataset as a graph~\cite{CaSn13}. A dense collection of multiple views allows 3D representations are possible, \eg structured from motion~\cite{IZFB09}, searching 2D features in 3D models~\cite{SaLK12,LSHF12}, or simultaneous visual localization and mapping~\cite{CuNe10}. However, this does not apply to sparse ``street-view'' imagery~\cite{TSP+13,TAS+15}, where dataset items and queries are groups of images taken at a single location, in a panorama-like layout.

Several approaches on visual instance retrieval propose to jointly represent a set of images.
These sets of images can appear at the query or at the database side. 
In the former case, these images are different views of the same object or scene~\cite{AZ12+,SJ15} and finally performance is improved. This joint representation, which commonly is an average query vector constructed via aggregation, is presumably more robust than each individual query vector. 
On the other hand, when aggregating images on the database side it is better to group them together by similarity~\cite{IFGRJ14}; images are assigned to sets, and a joint representation is created per set.

This work revisits location recognition by aggregating images both on query and database sides. 
Our method resembles implicit construction of a panorama, \ie images are combined in the feature space and not in the image space, but we also experiment with an explicit construction. Contrary to the general case of visual instance retrieval, it is easy to obtain multiple query images, e.g. capturing them with a smartphone or with multiple cameras in the case of autonomous driving. On the database side, location provides a natural way of grouping images together. Thus, contrary to generic retrieval, the images to be aggregated on the query and database sides, may not be similar to each other; they rather depict whatever is visible around a particular location.

We significantly outperform the state of the art without any form of supervision other than the natural, location-based grouping of images, and without any costly offline process like 3D reconstruction. Indeed we are reaching near perfect location recognition on the Pittsburgh dataset~\cite{TSP+13} even when we use as few as four views on the query side.

\begin{figure*}[t!]
\vspace{-10pt}
\begin{tabular}{ccc}
       % correspondence colors
\definecolor{r1}{RGB}{0,     0, 255}  % blue
\definecolor{r2}{RGB}{0,   128,   0}  % dark green
\definecolor{r3}{RGB}{128,   0,   0}  % brown
\definecolor{r5}{RGB}{0,   220, 220}  % cyan
\definecolor{r4}{RGB}{255,   0, 255}  % magenta
\definecolor{r0}{RGB}{220, 220,   0}  % yellow
\definecolor{r7}{RGB}{0,     0, 128}  % dark blue
\definecolor{r8}{RGB}{0,   128, 128}
\definecolor{r9}{RGB}{128,   0, 128}
\definecolor{r6}{RGB}{128, 128,   0}
% correspondence styles
\tikzset{corA/.style   = {fill=#1,circle,inner sep=1pt, draw=black, thick}}
\tikzset{corB/.style   = {fill=#1,rectangle,inner sep=2pt}}
%================================================================
%----------------------------------------------------------------
\raisebox{7pt}{
\extfig{toyLeft}{
\resizebox{4.0cm}{3.6cm}{%
\begin{tikzpicture}[scale=2, font=\footnotesize]
   % correspondence coordinates
   \def\corcoord{
      \coordinate (a1) at (1.2/5, 9.2/5);
      \coordinate (a2) at (2/5, 9.2/5);
      \coordinate (a3) at (2.8/5, 9.2/5);
      \coordinate (a4) at (9/5, 9/5);
      \coordinate (a5) at (5/5, 4.2/5);
      \coordinate (a6) at (5/5, 5/5);
      \coordinate (a7) at (5/5,5.8/5);
      \coordinate (a8) at (9.2/5,1/5);
      \coordinate (b1) at (2/5, 10/5);
      \coordinate (b2) at (5.8/5, 4.2/5);
      \coordinate (b3) at (5.8/5, 5/5);
      \coordinate (b4) at (5.8/5, 5.8/5);
      \coordinate (b5) at (10/5, 1/5);
      \coordinate (b6) at (1.2/5, 1/5);
      \coordinate (b7) at (2/5,1/5);
      \coordinate (b8) at (2/5,1.8/5); 
   }

   % correspondence nodes
   \def\cornodes{
      \node[corA=r1] (p1) at (a1) {\footnotesize \color{white} $x_1$};
      \node[corA=r1] (p2) at (a2) {\footnotesize \color{white} $x_2$};
      \node[corA=r1] (p3) at (a3) {\footnotesize \color{white} $x_3$};
      \node[corA=r1] (p4) at (a4) {\footnotesize \color{white} $x_4$};
      \node[corA=r1] (p5) at (a5) {\footnotesize \color{white} $x_5$};
      \node[corA=r1] (p6) at (a6) {\footnotesize \color{white} $x_6$};
      \node[corA=r1] (p7) at (a7) {\footnotesize \color{white} $x_7$};
      \node[corA=r1] (p8) at (a8) {\footnotesize \color{white} $x_8$};
      
      \node[corA=r2] (p1) at (b1) {\footnotesize \color{white} $y_1$};
      \node[corA=r2] (p2) at (b2) {\footnotesize \color{white} $y_2$};
      \node[corA=r2] (p3) at (b3) {\footnotesize \color{white} $y_3$};
      \node[corA=r2] (p4) at (b4) {\footnotesize \color{white} $y_4$};
      \node[corA=r2] (p5) at (b5) {\footnotesize \color{white} $y_5$};
      \node[corA=r2] (p6) at (b6) {\footnotesize \color{white} $y_6$};
      \node[corA=r2] (p7) at (b7) {\footnotesize \color{white} $y_7$};
      \node[corA=r2] (p8) at (b8) {\footnotesize \color{white} $y_8$}; 
   }

   % level 3
   \begin{scope}[xshift=2.2cm]
      \draw (0,0) rectangle (2.2,2.2);
      \corcoord \cornodes
   \end{scope}

\end{tikzpicture}
}
}
}

&

\pgfplotsset{
    matrix plot/.style={
        tick label style  = {font=\footnotesize},
		axis on top,
        clip marker paths=true,
        scale only axis,
        height=\matrixrows/\matrixcols*\pgfkeysvalueof{/pgfplots/width},
        enlarge x limits={rel=0.5/\matrixcols},
        enlarge y limits={rel=0.5/\matrixrows},
        scatter/use mapped color={draw=mapped color, fill=mapped color},
        scatter,
        point meta=explicit,
        mark=square*,
        cycle list={
        mark size=0.5*\pgfkeysvalueof{/pgfplots/width}/\matrixcols
        }
    },
    matrix rows/.store in=\matrixrows,
    matrix rows=8,
    matrix cols/.store in=\matrixcols,
    matrix cols=8
}
%\extfig{toyMiddle}{

  \begin{tikzpicture}
    \begin{axis}[
            width=3.4cm, 
            matrix plot,
            colormap/hot,
%             colorbar,
			xticklabels={$x_1$, $x_2$, $x_3$, $x_4$, $x_5$, $x_6$, $x_7$, $x_8$},
            xtick={1,...,8},
            yticklabels={$y_1$, $y_2$, $y_3$, $y_4$, $y_5$, $y_6$, $y_7$, $y_8$},
            ytick={1,...,8},
            point meta max=1,
            point meta min=-0.5
        ]
      \addplot table [meta=funceval] {
x y funceval
1 1 0.278
2 1 0.527
3 1 0.278
4 1 0.000
5 1 0.000
6 1 0.000
7 1 0.000
8 1 0.000
1 2 0.000
2 2 0.000
3 2 0.000
4 2 0.000
5 2 0.527
6 2 0.278
7 2 0.041
8 2 0.000
1 3 0.000
2 3 0.000
3 3 0.000
4 3 0.000
5 3 0.278
6 3 0.527
7 3 0.278
8 3 0.000
1 4 0.000
2 4 0.000
3 4 0.000
4 4 0.000
5 4 0.041
6 4 0.278
7 4 0.527
8 4 0.000
1 5 0.000
2 5 0.000
3 5 0.000
4 5 0.000
5 5 0.000
6 5 0.000
7 5 0.000
8 5 0.527
1 6 0.000
2 6 0.000
3 6 0.000
4 6 0.000
5 6 0.000
6 6 0.000
7 6 0.000
8 6 0.000
1 7 0.000
2 7 0.000
3 7 0.000
4 7 0.000
5 7 0.000
6 7 0.000
7 7 0.000
8 7 0.000
1 8 0.000
2 8 0.000
3 8 0.000
4 8 0.000
5 8 0.000
6 8 0.000
7 8 0.000
8 8 0.000
      };
    \end{axis}    
  \end{tikzpicture}
%}

& 

\pgfplotsset{
    matrix plot/.style={
        tick label style  = {font=\footnotesize},
		axis on top,
        clip marker paths=true,
        scale only axis,
        height=\matrixrows/\matrixcols*\pgfkeysvalueof{/pgfplots/width},
        enlarge x limits={rel=0.5/\matrixcols},
        enlarge y limits={rel=0.5/\matrixrows},
        scatter/use mapped color={draw=mapped color, fill=mapped color},
        scatter,
        point meta=explicit,
        mark=square*,
        cycle list={
        mark size=0.5*\pgfkeysvalueof{/pgfplots/width}/\matrixcols
        }
    },
    matrix rows/.store in=\matrixrows,
    matrix rows=8,
    matrix cols/.store in=\matrixcols,
    matrix cols=8
}
%\extfig{toyRight}{
  \begin{tikzpicture}
    \begin{axis}[
            width=3.4cm, 
            matrix plot,
            colormap/hot,
            colorbar,
	    	colorbar style={},  
			xticklabels={$x_1$, $x_2$, $x_3$, $x_4$, $x_5$, $x_6$, $x_7$, $x_8$},
            xtick={1,...,8},
            yticklabels={$y_1$, $y_2$, $y_3$, $y_4$, $y_5$, $y_6$, $y_7$, $y_8$},
            ytick={1,...,8},
            point meta max=1,
            point meta min=-0.5
        ]
      \addplot table [meta=funceval] {
x y funceval
1 1 0.000
2 1 0.527
3 1 0.000
4 1 0.000
5 1 -0.000
6 1 0.000
7 1 -0.000
8 1 0.000
1 2 -0.000
2 2 0.000
3 2 -0.000
4 2 0.000
5 2 0.792
6 2 -0.533
7 2 0.220
8 2 0.000
1 3 0.000
2 3 -0.000
3 3 0.000
4 3 -0.000
5 3 -0.533
6 3 1.090
7 3 -0.533
8 3 -0.000
1 4 -0.000
2 4 0.000
3 4 -0.000
4 4 0.000
5 4 0.220
6 4 -0.533
7 4 0.792
8 4 0.000
1 5 0.000
2 5 -0.000
3 5 0.000
4 5 -0.000
5 5 -0.000
6 5 0.000
7 5 -0.000
8 5 0.527
1 6 0.000
2 6 -0.000
3 6 0.000
4 6 0.000
5 6 -0.000
6 6 0.000
7 6 -0.000
8 6 0.000
1 7 0.000
2 7 -0.000
3 7 0.000
4 7 0.000
5 7 -0.000
6 7 0.000
7 7 -0.000
8 7 0.000
1 8 -0.000
2 8 0.000
3 8 -0.000
4 8 -0.000
5 8 0.000
6 8 -0.000
7 8 0.000
8 8 -0.000

      };
    \end{axis}    
  \end{tikzpicture}
\\
%}

\end{tabular}
\vspace{-15pt}
\caption{Left: Toy example of two vector sets $\X$, $\Y$ on the 2D plane are shown on the left.
Middle: Pairwise similarity between all vectors, cross-matching with sum-vectors, \ie $\X^\top \Y$~\eqref{equ:sumToSum}. Only for visualization purposes, and since we are dealing with unnormalized 2D vectors, the similarity between vectors $\x,\y$ is defined as $e^{\|\x-\y\|^2}$.
Right: weighted pairwise-similarity between all vectors, cross-matching with pinv-vectors, \ie $\G_\X^{-1} \X^{\top} \Y \G_\Y^{-1}$~\eqref{equ:pinvToPinv}.
\label{fig:toyFig}
\vspace{-10pt}
}
\end{figure*}
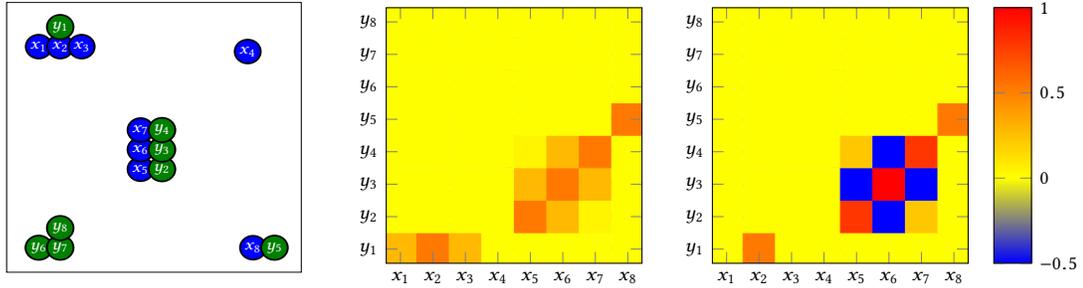
\section{Background}
This section describes the related work on Convolutional Neural Network (CNN) based descriptors for image retrieval and on image set joint representations. Our approach applies these methods on the dataset and query images.

\subsection{CNN Descriptors for Retrieval}
CNN-based global descriptors are becoming popular in image retrieval, especially for instance-level search.
Existing works~\cite{BL15,RSAC14,KMO15,TSJ15} employ ``off-the-shelf'' networks, originally trained on ImageNet, to extract descriptors via various pooling strategies. This offers invariance to geometric transformation and robustness to background clutter.
Other approaches~\cite{BSCL14,RTC16,GARL16} fine-tune such networks to obtain descriptor representations specifically adapted for instance search.

NetVLAD~\cite{AGT+15} is a recent work that trains a VLAD layer on top of convolutional layers in an end-to-end manner. It is tuned for the location recognition task.
The training images are obtained from panoramas, fed to a triplet loss to make it more compatible with image retrieval.
As a result, their representation outperforms existing works in standard location recognition benchmarks.

\subsection{Representing Sets of Vectors}
\label{subsec:agg}
Two common scenarios aggregate a set of vectors into a single vector representation for image retrieval.
The first case involves aggregation of a large number of local descriptors, either to reduce the number of descriptors~\cite{TAJ13,ShAJ15}, or to create a global descriptor~\cite{GMJP14,JZ14}.
In the other case, which is exploited in this work, a set of global image descriptors is aggregated into a single vector representation to construct a joint representation for a set of images~\cite{IFGRJ14}.

In particular, we follow the two \emph{memory vector} construction strategies proposed by Iscen \etal~\cite{IFGRJ14}.
The first method simply computes the sum of all vectors in a set.
Given a set of vectors represented as the columns of a $d \times n$ matrix $\X = [\x_1,\ldots,\x_n] $ with $\x_i \in \real^d$, the \emph{sum} memory vector is defined as
%
% \vspace{-3pt}
\begin{equation}
	m(\X) =  \X \1_n.
\label{equ:sumMemVec}
\end{equation}
% \vspace{-3pt}
Assuming linearly independent columns ($n < d$), the second method is based on the Moore-Penrose pseudo-inverse $\X^+$~\cite{RM72}, given by
%
% \vspace{-3pt}
\begin{equation}
	m^+(\X) =  (\X^+)^\top \1_n = \X (\X^{\top}\X)^{-1} \1_n.
\label{equ:pinvMemVec}
\end{equation}
% \vspace{-3pt}
It is theoretically optimized for high dimensional spaces and performs better in practice. This paper refers to the sum memory vector~\eqref{equ:sumMemVec} as \emph{sum-vector}, and to the pseudo-inverse memory vector~\eqref{equ:pinvMemVec} as \emph{pinv-vector}.

\mypar{Aggregating Dataset Images.}
The main purpose of aggregating dataset images is to reduce the computation cost of similarity search at query time~\cite{IFGRJ14}.
Dataset vectors are assigned to sets in an off-line process, and each set is represented by a single (memory) vector.
At query time, the similarity between the query vector and each memory vector is computed, and memory vectors are ranked accordingly.
Then the query is only compared to the database vectors belonging to the top ranked sets.
This strategy eliminates the exhaustive computation of the similarities query vs. dataset vectors.
Existing works use random assignments to create the sets, or weakly-supervised assignment based on k-means or kd-tree~\cite{IAF16,IFGRJ14}.

\mypar{Aggregating Query Images.}
Aggregation of query images has been also studied for instance-level object retrieval.
Multiple images depicting the query object allow to better handle the problems of occlusion, view-point change, scale change and other variations.
Arandjelovic \etal~\cite{AZ12+} investigate various scenarios, such as average or max pooling on query vectors and creating SVM models.
Recently, Sicre and J{\'e}gou~\cite{SJ15} have shown that aggregating query vectors with pinv-vector improves the search quality.

Aggregation of dataset images offers speed and memory improvements at the cost of performance loss.
On the other hand, aggregation of query images is only applicable in the particular case of multiple available query images and offers performance improvements at no extra cost.
Our approach adopts aggregation on both sides for the first time while enjoying speed, memory and performance improvements.

\begin{figure*}
\begin{tabular}
   {*{12}{@{\ssp}c@{\ssp}}}

\includegraphics[width= 0.07\textwidth]{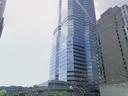} &
\includegraphics[width= 0.07\textwidth]{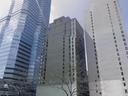} &
\includegraphics[width= 0.07\textwidth]{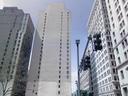} &
\includegraphics[width= 0.07\textwidth]{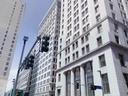} &
\includegraphics[width= 0.07\textwidth]{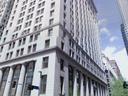} &
\includegraphics[width= 0.07\textwidth]{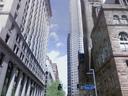} &
\includegraphics[width= 0.07\textwidth]{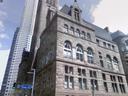} &
\includegraphics[width= 0.07\textwidth]{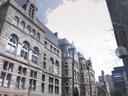} &
\includegraphics[width= 0.07\textwidth]{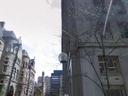} &
\includegraphics[width= 0.07\textwidth]{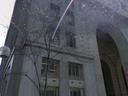} &
\includegraphics[width= 0.07\textwidth]{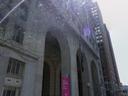} &
\includegraphics[width= 0.07\textwidth]{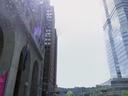} \\

\includegraphics[width= 0.07\textwidth]{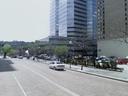} &
\includegraphics[width= 0.07\textwidth]{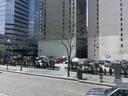} &
\includegraphics[width= 0.07\textwidth]{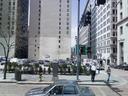} &
\includegraphics[width= 0.07\textwidth]{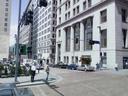} &
\includegraphics[width= 0.07\textwidth]{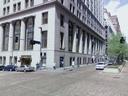} &
\includegraphics[width= 0.07\textwidth]{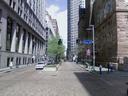} &
\includegraphics[width= 0.07\textwidth]{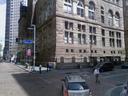} &
\includegraphics[width= 0.07\textwidth]{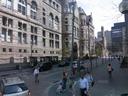} &
\includegraphics[width= 0.07\textwidth]{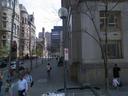} &
\includegraphics[width= 0.07\textwidth]{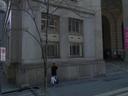} &
\includegraphics[width= 0.07\textwidth]{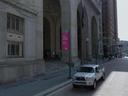} &
\includegraphics[width= 0.07\textwidth]{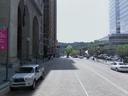} \\

\multicolumn{12}{c}{\includegraphics[width=0.9\textwidth]{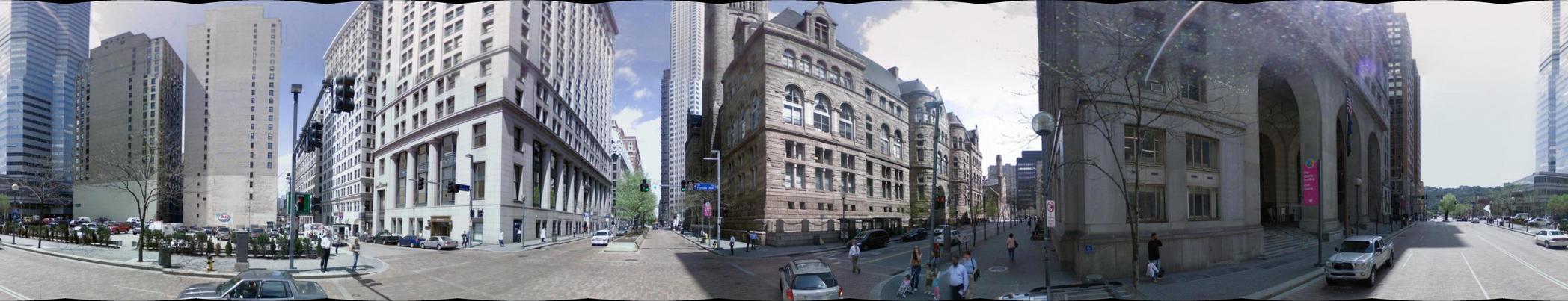}}

\end{tabular}
\vspace{-8pt}
\caption{Example of all images assigned to a single location (first two rows) and the corresponding panorama (last row) covering a full 360 degree view, constructed by automatic stitching.
\label{fig:panorama}
\vspace{-8pt}
}
\end{figure*}

\section{Panorama to Panorama Matching}
\label{sec:method}
This section describes our contribution for location recognition.
We assume that for each possible location we are given a set of images covering a full 360 degree view while consecutive images have an overlap (see Figure~\ref{fig:panorama}).
We propose two ways to construct a \emph{panoramic representation} of each location: an implicit way by vector aggregation and an explicit way by image stitching into a panorama and extraction of a single descriptor.

\subsection{Implicit Panorama Construction}
\label{subsec:pAgg}
We form a panoramic representation by aggregating the descriptors of images from the same location.
In this way, we implicitly construct a panorama in the descriptor space.
In order to achieve this, we employ two approaches for creating memory vectors, \ie sum-vector~\eqref{equ:sumMemVec} and pinv-vector~\eqref{equ:pinvMemVec}.

In contrast to previous works that aggregate the image vectors only on the dataset side~\cite{IFGRJ14} or only on the query side~\cite{SJ15}, we rather do it for both.
This requires that the query is also defined by a set of images which offer a 360 degree view. A realistic scenario of this context is autonomous driving and auto-localization where the query is defined by such a set of images.

Assume that $n$ images in a dataset location are represented by $d \times n$ matrix $\X$ and that $k$ images in the query location by $d \times k$ matrix $\Y$.
Analyzing the similarity between the two sum-vectors is straightforward.
Panorama similarity is given by the inner product
\begin{equation}
\vspace{-3pt}
	s(\X,\Y) = m(\X)^\top m(\Y) = \1_n^{\top} \X^{\top} \Y \1_k.
\label{equ:sumToSum}
\vspace{-2pt}
\end{equation}
Similarly, panorama similarity for pinv-vectors is given by
\begin{equation}
% \vspace{-2pt}
s^+(\X,\Y) = m^+(\X)^\top m^+(\Y) = \1_n^{\top} \G_\X^{-1} \X^{\top} \Y \G_\Y^{-1} \1_k,
\label{equ:pinvToPinv}
\end{equation}
% \vspace{-3pt}
%
where $\G_\X = \X^{\top}\X$ is the Gram matrix for $\X$.
Compared to~\eqref{equ:sumToSum}, the sum after cross-matching is weighted now, and the weights are given by $\G_\X^{-1}$ and $\G_\Y^{-1}$.
This is interpreted as ``democratizing'' the result of cross-matching; the contribution of vectors that are similar within the same set are down-weighted, just as in handling the burstiness phenomenon for local descriptors~\cite{JZ14}.
We visualize this with a toy example in Figure~\ref{fig:toyFig}. Unweighted cross-matching is dominated by ``bursty'' vectors in the same cluster. Democratization down-weights these contributions.

\subsection{Explicit Panorama Construction}
\label{subsec:netvladAgg}
Our second approach explicitly creates a panoramic image. The descriptors are then extracted from the panorama.
Given that images of a location are overlapping, we construct a panoramic image using an existing stitching method.
In particular, we use the work of Brown and Lowe~\cite{BL07}, which aligns, stitches, and blends images automatically based on their local SIFT descriptors and inlier correspondences.
Figure~\ref{fig:panorama} shows a stitched panoramic image.
Once stitching is complete, we extract a single global descriptor from the panorama image, capturing the entire scene.

\section{Experiments}
In this section, we describe our experimental setup, and compare our method to a number of baselines using the state-of-the-art NetVLAD network in a popular location recognition benchmark.

\begin{figure}
\begin{tabular}
   {*{2}{@{\ssp}c@{\ssp}}}
%\hline
 Query:  &
\includegraphics[width= 0.37\textwidth]{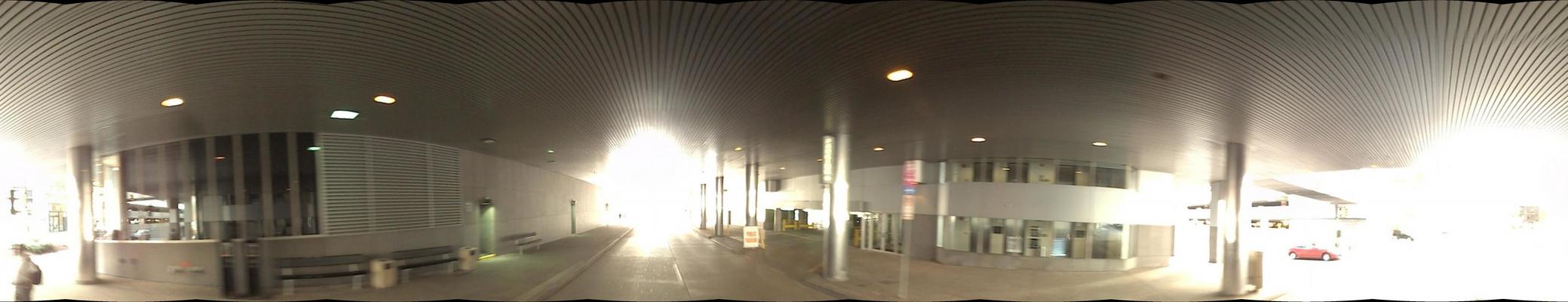} \\
Rank-1: &
\includegraphics[width= 0.37\textwidth]{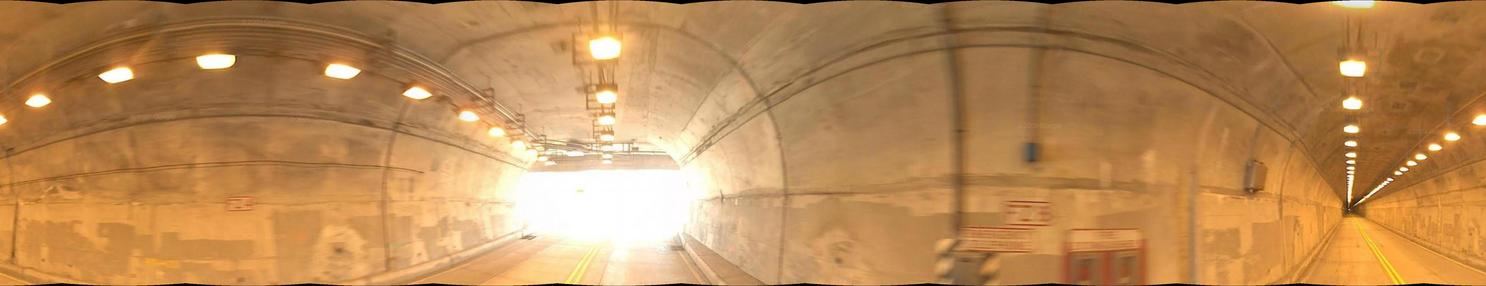} \\
%\hline

%\hline
Query: &
\includegraphics[width= 0.37\textwidth]{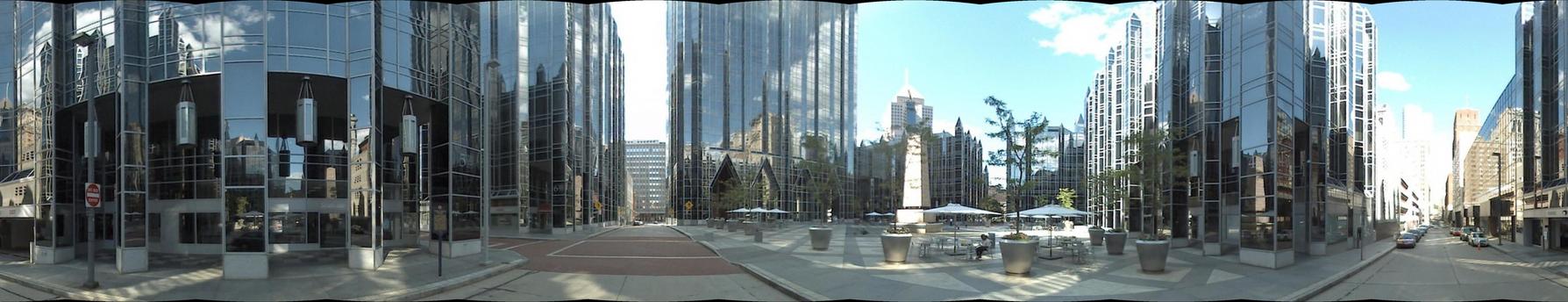} \\
Rank-1: &
\includegraphics[width= 0.37\textwidth]{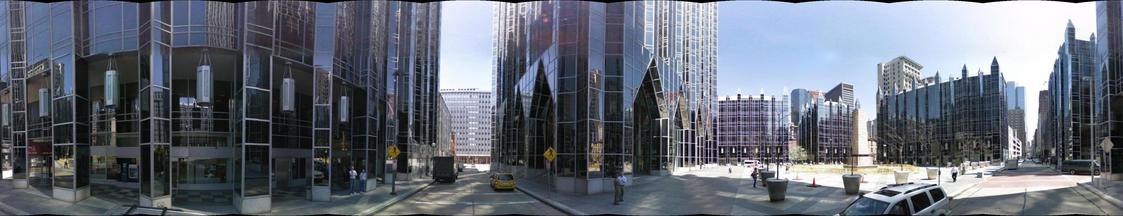} \\
%\hline

\end{tabular}
\vspace{-10pt}
\caption{Two examples of failures with  \emph{\ppn}. We show the query and the top ranked image from the dataset.
\label{fig:fail}
\vspace{-10pt}
}
\end{figure}

\begin{figure*}
\vspace{-8pt}
\centering
\input{figs/data/sample}
\begin{tabular}{cc}
\extfig{compFull}{
\begin{tikzpicture}
\begin{axis}[%
	width=0.48\linewidth,
	height=0.25\linewidth,
	xlabel={$N$},
	ylabel={recall@$N$},
	legend pos=south east,
    legend style={cells={anchor=east}, font =\footnotesize, fill opacity=0.8, row sep=-2.5pt},
    ymax = 100,
    ymin = 60,
    xmin = 1,
    xmax = 20,
    grid=both,
    xtick={1,5,10,20},
 	 y label style={at={(axis description cs:0.05,.5)}},
 	 x label style={at={(axis description cs:.5,.05)}}
]
	\addplot[color=black,    solid, mark=*,  mark size=1.5, line width=1.0] table[x=N, y expr={100*\thisrow{baseline}}]  \compFull;\leg{\ii};
	\addplot[color=magenta,     solid, mark=*,  mark size=1.5, line width=1.0] table[x=N, y expr={100*\thisrow{pinvToDb}}]  \compFull;\leg{\pi};
	\addplot[color=orange,    solid, mark=*,  mark size=1.5, line width=1.0] table[x=N, y expr={100*\thisrow{qToPinv}}]  \compFull;\leg{\ip};
	\addplot[color=brown,     solid, mark=*,  mark size=1.5, line width=1.0] table[x=N, y expr={100*\thisrow{sumToSum}}]  \compFull;\leg{\pps};
	\addplot[color=red,     solid, mark=*,  mark size=1.5, line width=1.0] table[x=N, y expr={100*\thisrow{pinvToPinv}}]  \compFull;\leg{\ppp};
	\addplot[color=blue,     solid, mark=*,  mark size=1.5, line width=1.0] table[x=N, y expr={100*\thisrow{NetVlad}}]  \compFull;\leg{\ppn};

\end{axis}
\end{tikzpicture}
}
&
\extfig{compSmall}{
\begin{tikzpicture}
\begin{axis}[%
	width=0.48\linewidth,
	height=0.25\linewidth,
	xlabel={$N$},
	ylabel={recall@$N$},
	legend pos=south east,
    legend style={cells={anchor=east}, font =\footnotesize, fill opacity=0.8, row sep=-2.5pt},
    ymax = 100,
    ymin = 30,
    xmin = 1,
    xmax = 20,
    grid=both,
    xtick={1,5,10,20},
 	 y label style={at={(axis description cs:0.05,.5)}},
 	 x label style={at={(axis description cs:.5,.05)}}
]
	\addplot[color=black,    solid, mark=*,  mark size=1.5, line width=1.0] table[x=N, y expr={100*\thisrow{baseline}}]  \compSmall;\leg{\ii};
	\addplot[color=magenta,     solid, mark=*,  mark size=1.5, line width=1.0] table[x=N, y expr={100*\thisrow{pinvToDb}}]  \compSmall;\leg{\pi};
	\addplot[color=orange,    solid, mark=*,  mark size=1.5, line width=1.0] table[x=N, y expr={100*\thisrow{qToPinv}}]  \compSmall;\leg{\ip};
	\addplot[color=brown,     solid, mark=*,  mark size=1.5, line width=1.0] table[x=N, y expr={100*\thisrow{sumToSum}}]  \compSmall;\leg{\pps};
	\addplot[color=red,     solid, mark=*,  mark size=1.5, line width=1.0] table[x=N, y expr={100*\thisrow{pinvToPinv}}]  \compSmall;\leg{\ppp};
	\addplot[color=blue,     solid, mark=*,  mark size=1.5, line width=1.0] table[x=N, y expr={100*\thisrow{NetVlad}}]  \compSmall;\leg{\ppn};

\end{axis}
\end{tikzpicture}
}
\end{tabular}
\vspace{-8pt}
\vspace{-10pt}
\caption{Comparison of existing approaches (im2im~\cite{AGT+15}, im2pan~\cite{IFGRJ14}, pan2im~\cite{SJ15}) with our methods (pan2pan/sum, pan2pan/pinv and pan2pan/net) for the full 4096D (left) and for reduced dimensionality to 256D (right).
\vspace{-8pt}
\label{fig:compFig}
}
\end{figure*}

\begin{figure}[t]
\centering
\input{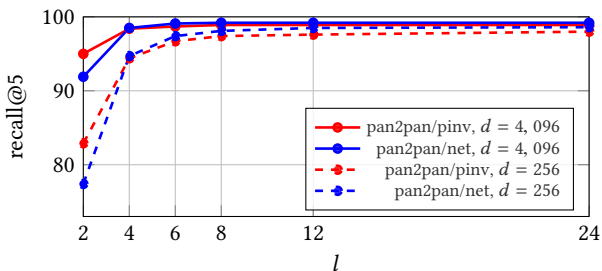}
\begin{tabular}{c}
\extfig{sampleAll}{
\begin{tikzpicture}
\begin{axis}[%
	width=0.98\linewidth,
	height=0.5\linewidth,
	xlabel={$l$},
	ylabel={recall@5},
	title={},
	legend pos=south east,
    legend style={cells={anchor=east}, font =\footnotesize, fill opacity=0.8, row sep=-2.5pt},
    ymax = 100,
    ymin = 73,
    xmin = 2,
    xmax = 24,
    grid=both,
    xtick={2,4,6,8,12,24},
 	 y label style={at={(axis description cs:0.05,.5)}},
 	 x label style={at={(axis description cs:.5,.05)}}
]
	\addplot[color=red,     solid, mark=*,  mark size=1.5, line width=1.0] table[x=l, y expr={100*\thisrow{pinvToPinv}}]  \sampleAll;\leg{\ppp, $d = 4,096$};
	\addplot[color=blue,     solid, mark=*,  mark size=1.5, line width=1.0] table[x=l, y expr={100*\thisrow{NetVlad}}]  \sampleAll;\leg{\ppn, $d = 4,096$};
	\addplot[color=red,     dashed, mark=*,  mark size=1.5, line width=1.0] table[x=l, y expr={100*\thisrow{pinvSmall}}]  \sampleAll;\leg{\ppp, $d = 256$};
	\addplot[color=blue,     dashed, mark=*,  mark size=1.5, line width=1.0] table[x=l, y expr={100*\thisrow{NetVladSmall}}]  \sampleAll;\leg{\ppn, $d = 256$};

\end{axis}
\end{tikzpicture}
}
\end{tabular}
\vspace{-8pt}
\vspace{-10pt}
\caption{Recall@5 on Pitt250k, sampling $l$ images from each query panorama and using NetVLAD descriptors of two different dimensionalities $d$. We report average measurements over 10 random experiments and compare our methods \ppp and \ppn.
\label{fig:sampleFig}
\vspace{-8pt}
}
\end{figure}

\subsection{Experimental Setup}
The methods are evaluated on the Pittsburgh dataset~\cite{TSP+13} referred to as Pitt250k.
It contains 250k database and 24k query images from Google Street View.
It is split into training, validation, and test sets~\cite{AGT+15}.
We evaluate our approach on the test set, which consists of 83,952 dataset images and 8,280 query images.
Each image is associated with a GPS location and 24 images are associated with the same GPS location.
Therefore, each panoramic representation aggregates 24 images.
There is a total of 345 query locations and 3,498 dataset locations.
We use NetVLAD for our descriptor representation in all experiments.
While the original representation is $d=$ 4,096 dimensional, we also experiment with reducing dimensionality to $d=256$ by PCA.

The standard evaluation metric is Recall@$N$.
It is defined to equal  1 if at least one of the top $N$ retrieved dataset images is within 25 meters from the spatial location of the query.
Average is reported over all queries.
We follow this protocol for the baseline and other cases where the query images are used individually.

Aggregating on the query side implies that there is a single query per location: the number of queries decreases from 8,280 to 345.
We report the average recall@$N$ from these 345 panorama queries. Section~\ref{sec:sparse} also experiments with a larger number of random queries, each capturing only a fraction of the panoramic view. In this case, recall@$N$ is averaged over those random queries.
Aggregating on the dataset side does not affect the standard evaluation.

\subsection{Panorama Matching}
We refer to our proposed method as panorama to panorama or \emph{\pp} matching, in particular \emph{\pps} and \emph{\ppp} when aggregating descriptors with sum-vector and pinv-vector respectively; and as \emph{\ppn} when using a NetVLAD descriptor from an explicit panorama. We compare against the following baselines:
image to image matching (\emph{\ii}) as in the work by Arandjelovic \etal~\cite{AGT+15},
image to panorama matching (\emph{\ip}) corresponding to dataset-side aggregation as in the work by Iscen et al.~\cite{IFGRJ14},
and panorama to image matching (\emph{\pi}) corresponding to query-side aggregation as in the work by Sicre and J\'egou~\cite{SJ15}.

Figure~\ref{fig:compFig} compares all methods for different descriptor dimensions.
Clearly, panorama to panorama matching outperforms all other methods.
The improvement is consistent for all $N$ and significant for low $N$: \ppn obtains 98\% recall@1!
There are only 7 failure queries. 
Two of them are shown in Figure~\ref{fig:fail}. One is a challenging query depicting an indoor parking lot and the other actually retrieves the same building, which is incorrectly marked in the dataset's ground truth.

The recall is not only improved, but the search is also more efficient both speed-wise ($24^2 \times$ faster) and memory-wise ($24 \times$ less memory).
Instead of comparing a given query image against 83k vectors, we only make 3.5k comparisons.
Additional operations are introduced when aggregating the set of query images, but this cost is fixed and small compared to the savings from the dataset side.

Comparing to results in prior work, \ip behaves as in the work of Iscen \etal~\cite{IFGRJ14} when compared to the baseline \ii. That is, memory compression and speed up at the cost of reduced performance. However, \pi does not appear to be effective in our case, in contrast to the work of Sicre and J\'egou~\cite{SJ15}. On the contrary, \pp significantly improves the performance while enjoying both memory compression and search efficiency.

\subsection{Sparse Panorama Matching}
\label{sec:sparse}
Aggregating on the dataset side is performed off-line.
However, the user is required to capture images and to construct a full panorama (24 images in our case) at query time.
Even though this is not a daunting task given the advances of smartphones and tablets, we additionally investigate a scenario where the user only captures a partial panoramic view.

In particular, we randomly sample a subset of $l$ images from the query location and consider them as the query image set. Explicit panorama construction is no longer possible because the sampled images may not overlap and so we cannot stitch them. In this case, we feed sampled images through the convolutional layers only, and stack together all activations before pooling them through the NetVLAD layer (\ppn for sparse panoramas).

Figure~\ref{fig:sampleFig} shows the results.
Our methods have near-perfect performance even for a small number of sampled images.
When the user only takes four random photos, we are able to locate them up to ~99\% recall@5. Another interesting observation is that \ppp outperforms \ppn for $l=2$, which is expected due to the nature of pinv-vec construction.
It is theoretically shown to perform well even if all the vectors in the set are random, as shown in the original paper~\cite{IFGRJ14}.

\subsection{Comparison to Diffusion-based Retrieval}
This work casts location recognition as a retrieval task. Query expansion techniques significantly improve retrieval performance.
We compare to the state-of-the-art retrieval method by Iscen \etal~\cite{ITA+16}, a kind of query expansion based on graph diffusion.
In this method, an image is represented by individual region descriptors and at query time all query regions are processed.
We compare to this method by considering that regions and images in~\cite{ITA+16} correspond to images and panoramas respectively in our scenario.

Our \ppp and \ppn approaches achieve 96.5\% and 98\% recall@1 respectively, while the approach~\cite{ITA+16} gives 91.9\%. Even though query expansion improves the baseline, it does not help as much as our methods.
This can be expected because~\cite{ITA+16} is based on many instances of the same object, which is not the case for location recognition on street view imagery.

\section{Conclusions}
Our method is unsupervised and conceptually very simple, yet very effective. Besides the performance gain, we make significant savings in space by aggregating descriptors of individual images over each group.
The need for multiple query views is not very demanding because only four views are enough---an entire query panorama is definitely not needed.

Although our aggregation methods have been used for instance retrieval in the past, we are the first to successfully aggregate on both dataset and query-side for location recognition (which in fact has failed for instance retrieval~\cite{ShAJ15}).
An interesting finding is that although the NetVLAD descriptor has been explicitly optimized to aggregate CNN activations on the location recognition task, in some cases it is preferable to aggregate individual views into a pinv-vector rather than extracting a single NetVLAD descriptor from an explicit panorama.

\textbf{Acknowledgments.}
The authors were supported by the MSMT LL1303 ERC-CZ grant. The Tesla K40 used for this research was donated by the NVIDIA Corporation.

\newpage
\renewcommand{\bibfont}{\normalsize}
\bibliographystyle{abbrv}
\bibliography{egbib}
\flushend
\end{document}